\documentclass[runningheads]{llncs}
\usepackage[T1]{fontenc}
\usepackage{graphicx,verbatim}

\usepackage{textcomp}

\usepackage{amsmath} 
\usepackage{amssymb}
\usepackage{multirow}
\usepackage{makecell}
\usepackage{booktabs}

\usepackage{graphicx}

\usepackage[colorlinks=true, linkcolor=black, citecolor=black, urlcolor=blue]{hyperref}

\usepackage{color}
\begin{document}

\title{
Multimodal Stepwise Clinically-Guided Attention Learning for Pathological Complete Response Prediction in Breast Cancer
}
\titlerunning{Multimodal Stepwise Clinically-Guided Attention Learning}
%\begin{comment}  %% Removed for anonymized MICCAI submission
\author{Alice Natalina Caragliano\inst{1} \and
Valerio Guarrasi\inst{1} \and
Michela Gravina\inst{2} \and
Carlo Sansone\inst{2} \and
Paolo Soda\inst{1,3}}
\authorrunning{A. Caragliano et al.}
\institute{Unit of Artificial Intelligence and Computer Systems, Department of Engineering, Università Campus Bio-Medico di Roma, Rome, Italy \\
\email{\{a.caragliano, valerio.guarrasi, p.soda\}@unicampus.it}\\
\and
Department of Electrical Engineering and Information Technology, \\ University of Naples Federico II, Naples, Italy\\
\email{\{michela.gravina, carlo.sansone\}@unina.it} \\ \and
Department of Diagnostics and Intervention, Radiation Physics, Biomedical Engineering, Umeå University, Umeå, Sweden 
\\ \email{\{paolo.soda\}@umu.se}
}
%\end{comment}
\begin{comment}
\author{Anonymized Authors}  %% Added for anonymized MICCAI submission
\authorrunning{A. Caragliano et al.}
\institute{Anonymized Affiliations \\
    \email{email@anonymized.com}}
\end{comment} 
\maketitle              % typeset the header of the contribution
\begin{abstract}
%The abstract should briefly summarize the contents of the paper in 150--250 words.  If you are to include a link to your Repository, please make sure it is anonymized for the double-blind review phase.

Pathological complete response (pCR) is a key prognostic factor in breast cancer patients undergoing neoadjuvant therapy, strongly associated with long-term survival and treatment personalization. However, accurate pre-treatment pCR prediction remains challenging due to severe class imbalance and limited generalizability across diverse clinical settings. In this work, we propose a multimodal stepwise clinically-guided attention learning framework for pCR prediction from breast magnetic resonance imaging (MRI), designed to address these limitations through medically grounded spatial guidance and multimodal integration. The approach follows a stepwise training strategy inspired by physician reasoning: the model first learns global discriminative imaging patterns, then attention mechanisms are introduced to constrain the network toward tumor regions, and finally clinical variables are integrated to refine decision-making. This guidance strategy encourages prioritization of task-relevant features, improving identification of responders despite their limited representation in the dataset. Moreover, grounding attention in anatomically consistent tumor regions reduces reliance on dataset-specific patterns, thereby enhancing cross-institutional generalization. The framework is evaluated through external validation across heterogeneous MRI cohorts. Compared to non-guided single-stage baselines, the proposed approach improves sensitivity while maintaining competitive specificity, and produces anatomically coherent attention maps that support interpretation of the model’s predictions. These findings highlight the potential of clinically-guided multimodal attention learning for robust and generalizable pCR prediction in breast cancer. 
%Code will be released upon acceptance.

% Authors must provide keywords and are not allowed to remove this Keyword section.

\end{abstract}

% CONTRIBUTION
\begin{comment}
The primary contribution of this work is a multimodal stepwise clinically-guided attention learning framework for pathological complete response prediction. By integrating tumor localization guidance and clinical information, the proposed strategy addresses the complexity of treatment response modeling, mitigating class imbalance and enhancing generalization across multi-institutional cohorts. 
\end{comment}
%
%
\section{Introduction}

Breast cancer remains one of the leading causes of cancer-related mortality among women worldwide~\cite{who}. In patients undergoing neoadjuvant therapy (NAT) ~\cite{van2007preoperative}, pathological complete response (pCR), defined as the absence of residual invasive tumor at surgery, is a clinically meaningful endpoint associated with improved long-term outcomes~\cite{cortazar2014pathological,spring2020pathologic}. Reliable pre-treatment pCR prediction is of high clinical relevance, as it provides insight into NAT efficacy, thereby enabling personalized treatment strategies (e.g., less extensive surgery in likely responders).
Dynamic contrast-enhanced magnetic resonance imaging (DCE-MRI) provides spatial and functional characterization of tumor vascularity and enhancement kinetics, closely related to treatment response~\cite{cheng2020diagnostic,kang2020evaluating}, making it a strong modality for modeling pCR prior to therapy initiation.

Recent deep learning (DL) approaches for pCR prediction from breast MRI have demonstrated promising performance
%, further enhanced through multimodal integration of clinical data
~\cite{dammu2023deep,yeon2025br,fridman2025breastdcedl,musah2025large,awwad2025can}. However, despite encouraging results,  several limitations hinder the adoption of these models in clinical practice. First, pCR prediction is inherently affected by \textit{class imbalance}, as responder cases typically represent a minority of treated patients. This imbalance may bias models toward majority-class patterns, leading to suboptimal sensitivity for pCR detection. Second, \textit{limited cross-dataset generalization} remains a major barrier, as models trained on a single cohort may learn correlations tied to specific imaging protocols or patient distributions, resulting in performance degradation under external evaluation. Third, many approaches provide \textit{limited insight }into their decision-making process, reducing transparency and clinical trust.
Most existing studies for pCR prediction rely on internal validation or limited cross-dataset evaluation~\cite{dammu2023deep,yeon2025br,fridman2025breastdcedl}. Although some works include external cohorts~\cite{awwad2025can,musah2025large}, they primarily emphasize predictive accuracy without explicitly addressing these limitations. 

To overcome these challenges, we propose a multimodal stepwise clinically-guided attention learning framework for pCR prediction from breast DCE-MRI. This approach progressively structures training by constraining attention toward tumor regions and integrating clinical data. This design promotes biologically grounded feature learning, improving identification of responders despite their
limited representation while reducing reliance on dataset-specific correlations.

The main contributions of this work are:
(i) a multimodal stepwise framework for pCR prediction that progressively integrates spatial guidance and clinical information within an attention-based architecture;
(ii) a clinically-guided attention mechanism that promotes tumor-relevant feature learning, validated through cross-dataset experiments demonstrating improved class imbalance mitigation and stronger cross-cohort generalization;
and (iii) a clearer understanding of the model’s decision process via anatomically coherent attention maps.

\section{Methods} \label{sec:Methods}
\begin{figure}[!hbt] 
    \centering 
    \includegraphics[width=0.8\textwidth]{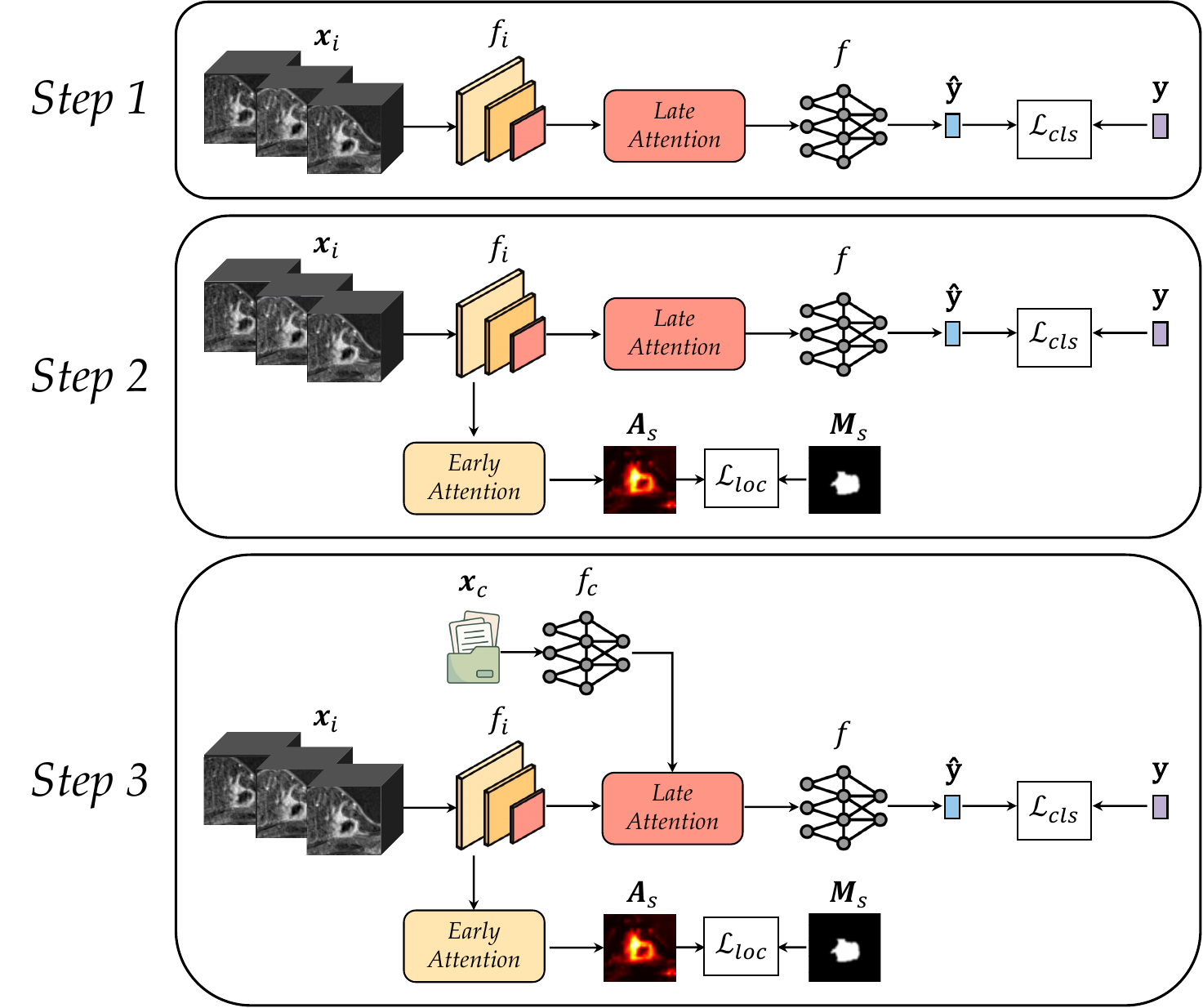} 
    \caption{{Overview of the proposed framework, structured into three progressive steps integrating attention mechanisms and clinically grounded guidance.}}
    \label{fig:method} 
\end{figure}

Since pCR prediction is closely associated with tumor morphology and clinical characteristics, we introduce medical knowledge into the training process via explicit tumor localization guidance to promote biologically grounded feature learning. Specifically, expert-provided tumor segmentations are used to guide the model’s attention, encouraging alignment between learned representations and tumor regions. The proposed framework ({Fig.~\ref{fig:method}}) progressively structures learning through spatial guidance and multimodal integration, ensuring focus on clinically relevant features. This strategy mirrors physician reasoning leveraging global image assessment, tumor-focused analysis, and integration of clinical information for final decision refinement.
\subsection{Problem Formulation}
Let $\mathbf{x}_i \in \mathbb{R}^{C \times D \times H \times W}$ denote a multi-phase 3D breast DCE-MRI input, where $C$ corresponds to post-contrast phases, each with depth D, height H, and width W. The corresponding label vector $\mathbf{y} = \{0,1\}$ denotes the binary pCR outcome, with $y = 1$ indicating responders.
For each input, a tumor segmentation mask $\mathbf{M} \in \{0,1\}^{1 \times D \times H \times W}$ is available and used during training to guide the learning process.
Moreover, each patient is associated with a clinical feature vector $\mathbf{x}_c \in \mathbb{R}^{K}$. Our objective is to accurately predict pCR by leveraging tumor-focused guidance to: (i) reduce majority-class bias, (ii) improve cross-dataset generalization, and (iii) support interpretation of the model’s decisions.

\subsection{Multimodal Stepwise Clinically-guided Attention Learning} \label{sec:method}
The proposed architecture is built upon a 3D transformer-based encoder $f_i$, coupled with an \textit{Early Attention} module operating on early-stage features during the spatially guided training steps, and a \textit{Late Attention} module, active throughout all steps to aggregate deep semantic representations for final classification ({Fig.~\ref{fig:method}}).
Let $\mathbf{F}_s \in \mathbb{R}^{d_s \times D_s \times H_s \times W_s}$ and $\mathbf{F}_l \in \mathbb{R}^{d_l \times D_l \times H_l \times W_l}$ denote the early-stage and deep semantic feature maps, respectively, where $d_s$ and $d_l$ represent the number of feature channels, and $(D_s, H_s, W_s)$ and $(D_l, H_l, W_l)$ their corresponding spatial resolution. A linear classification head $f$ maps the final representation produced by the \textit{Late Attention} module to pCR probability $\mathbf{\hat{y}}$. 

Our framework follows a stepwise learning paradigm that progressively introduces medically grounded constraints through three steps.
%(1) global imaging learning; (2) clinically-guided spatial attention learning; (3) multimodal clinically-modulated decision learning. 

\textit{\textbf{Step 1 - Global Imaging Learning.}}
This step focuses on learning only discriminative global imaging representations. The pCR prediction is obtained through the \textit{Late Attention} module, which operates on the deepest encoder representation $\mathbf{F}_l$. Given the input image $\mathbf{x}_{i}$, the encoder $f_{i}$ produces the feature map $\mathbf{F}_l$, which is reshaped into a sequence of tokens
$\mathbf{T}_l \in \mathbb{R}^{N_l \times d_l}$,
where $N_l = D_l \times H_l \times W_l$.
Each token encodes high-level semantic information from a spatial region of the volume, with $d_l$ representing the embedding dimension of the semantic representation.
A learnable class token $\mathbf{t}_{cls}^{(l)} \in \mathbb{R}^{1 \times d_l}$ is introduced to aggregate contextual information via multi-head attention, yielding the global representation: 
\begin{equation}
\mathbf{z}_{cls}^{(l)} = \mathrm{MHA}\big(\mathbf{t}_{cls}^{(l)}, \mathbf{T}_l, \mathbf{T}_l \big) \label{mha}
\end{equation}
where $\mathbf{z}_{cls}^{(l)} \in \mathbb{R}^{d_l}$ is subsequently mapped by $f$ to the predicted pCR probability.

\textit{\textbf{Step 2 - Clinically-Guided Spatial Attention Learning.}}
To constrain the model to focus on anatomically meaningful tumor regions, we introduce medically grounded spatial
guidance through the \textit{Early Attention} module operating on early encoder features. Since $\mathbf{F}_s$ preserves finer anatomical details compared to deeper semantic representations, it is well suited for spatial supervision: it is therefore reshaped into a sequence of tokens $\mathbf{T}_s \in \mathbb{R}^{N_s \times d_s}$, with $N_s = D_s \times H_s \times W_s$. A learnable class token $\mathbf{t}_{cls}^{(s)} \in \mathbb{R}^{1 \times d_s}$ is then used to aggregate contextual information from $\mathbf{T}_s$ through the multi-head attention mechanism defined in \eqref{mha}. 
The resulting attention weights across the $n_h$ attention heads, denoted as $\mathbf{A} \in \mathbb{R}^{n_h \times N_s}$, are averaged across heads and reshaped into an attention map $\mathbf{A}_{s} \in \mathbb{R}^{1 \times D \times H \times W}$, resampled to the input image resolution. 
During training, this map is aligned with the expert mask via a localization loss, encouraging focus on tumor regions. Final classification is performed through the \textit{Late Attention} block.  

%\begin{equation}
%\mathbf{A} = \mathrm{MHA}\big(\mathbf{t}_{cls}^{(s)}, %\mathbf{T}_s, \mathbf{T}_s \big),
%\end{equation}
%

\textit{\textbf{Step 3 - Multimodal Clinically-Modulated Decision Learning.}}
In the final step, clinical variables  $\mathbf{x}_c$ are integrated into the decision-making mechanism by modulating the \textit{Late Attention} block. Clinical features are projected into an embedding space through a linear layer $f_{c}$, yielding $\Delta_{c} \in \mathbb{R}^{d_l}$, which is added to $\mathbf{t}_{cls}^{(l)}$, resulting in a class token and global semantic aggregation which are influenced by clinical information. The same  multi-head attention mechanism defined in \eqref{mha} is adopted, yielding a final representation which is mapped by $f$ to produce the final pCR probability.    
%\begin{equation}
%$\mathbf{t}_{cls}^{(l*)} = \mathbf{t}_{cls}^{(l)} + \Delta_{c}$
%\end{equation}
%This formulation allows clinical information to influence global semantic aggregation, leveraging the same
%The resulting representation 
Importantly, the \textit{Early Attention} block remains active, preserving spatial guidance, while the \textit{Late Attention} becomes clinically aware. This step therefore enables multimodal decision integration while preserving the tumor-focused spatial alignment learned in the previous step.

To jointly optimize predictive performance and anatomically consistent attention alignment, the model is trained using a composite objective $\mathcal{L}$ combining classification $\mathcal{L}_{\mathit{cls}}$ and localization $\mathcal{L}_{\mathit{loc}}$ losses, defined as 
\begin{equation}
\mathcal{L} = \mathcal{L}_{\mathit{cls}} + \lambda \mathcal{L}_{\mathit{loc}} \label{loss}
\end{equation}
where $\lambda$ controls the strength of spatial guidance. When $\lambda= 0$, the model is optimized using $\mathcal{L}_{\mathit{cls}}$, defined as cross-entropy, while $\mathcal{L}_{\mathit{loc}}$ is applied when $\lambda > 0$
to encourage the model to align its focus with the expert-provided tumor mask, promoting clinically meaningful representations. To account for annotation uncertainty, we first apply a Gaussian smoothing 
%(kernel size = 5, $\sigma$ = 1.1) 
to the mask and then we define $\mathcal{L}_{\mathit{loc}}$ as the mean squared error between the attention map $\mathbf{A}_{s}$ and the blurred mask $\mathbf{M}_{s}$. 
\textit{Step 1} is trained using $\mathcal{L}_{\mathit{cls}}$, while $\mathcal{L}$ is employed in \textit{Step 2} and \textit{Step 3}. The three steps are trained sequentially, with each step initialized from the best-performing validation model of the previous step. All parameters are fine-tuned, and convergence at each step is determined via early stopping on validation loss. 

\section{Materials}
The proposed framework was evaluated on the MAMA-MIA dataset~\cite{garrucho2025}, comprising pre-treatment DCE-MRI scans of breast cancer patients undergoing NAT. The dataset aggregates four publicly available collections: DUKE~\cite{duke}, I-SPY1~\cite{ispy1}, I-SPY2~\cite{ispy2}, and NACT~\cite{nact}. Imaging protocols include both axial and sagittal acquisitions, as well as unilateral and bilateral examinations.
Given its multi-center composition, the MAMA-MIA dataset is particularly suitable for evaluating cross-dataset generalization. 
It includes 1,506 T1-weighted DCE-MRI studies with corresponding tumor segmentations.
After excluding cases without pCR labels and fat-suppressed sequences, the final cohort included 1,485 patients, of whom 437 (29.4\%)  with pCR.  The final cohort distribution was: DUKE (279 patients, 64 pCR), I-SPY1 (166 patients, 49 pCR), I-SPY2 (976 patients, 313 pCR), and NACT (64 patients, 11 pCR).   
Clinical variables are also provided, including age, multifocal disease status, hormone receptor status, human epidermal growth factor receptor 2 status, molecular subtype, and ethnicity.

\textit{Pre-processing.} Given the temporal heterogeneity of DCE-MRI protocols, which include a variable number of dynamic volumes, we standardize inputs by selecting three post-contrast phases per patient, corresponding to early enhancement, peak enhancement, and washout phases of the time-intensity curves.
To mitigate inter-dataset variability, a standardized pre-processing pipeline was applied. First, subtraction volumes were computed to enhance tumor evidence and suppress background enhancement. All sagittal acquisitions were reoriented to the axial plane to ensure consistent anatomical alignment, and bilateral examinations were converted to unilateral volumes using tumor masks to isolate the affected breast. 
N4 bias field correction was then applied to reduce intensity inhomogeneity artifacts. Subsequently, all volumes were resampled to a uniform voxel spacing of  0.7 $\times$ 0.7 $\times$ 2.0~$mm^{3}$, and cropped to a fixed size of $96 \times 96 \times 32$ voxels.
Finally, voxel intensities were clipped within a percentile-based range (1st–99th percentiles),
and linearly normalized into the [0,1] interval. 
The clinical variables were processed using a dedicated pipeline,
in which categorical variables (e.g., molecular subtype) were one-hot encoded,  
while the numerical feature (i.e., patient age) was standardized using z-score normalization. 

\section{Experimental Configuration}
To comprehensively evaluate the proposed framework, we design experiments along two complementary axes:
(i) architectural configurations, including stepwise evaluation and ablation studies; and
(ii) cross-dataset evaluation.

\textit{\textbf{Architectural Configurations and Ablation Studies.}}
We evaluate our framework by assessing the contribution of each step in the progressive training strategy (\hyperref[sec:method]{Section~\ref{sec:method}}). The intermediate stages naturally define architectural ablations: 
\textit{Step 1} corresponds to global imaging learning without spatial guidance, \textit{Step 2} introduces localization-guided attention without clinical modulation, and \textit{Step 3} represents the multimodal clinically-guided model. 

Additionally, to assess the independent contribution of each modality, we evaluate \textit{unimodal baselines}. The unimodal imaging configuration corresponds to the first two steps of our approach, while the unimodal clinical model ($CL$) is implemented through a multilayer perceptron (MLP) trained on clinical data. 

Moreover, we assess the contribution of the proposed \textit{multimodal fusion strategy}, by comparing it against conventional fusion approaches~\cite{ramachandram2017deep}. In \textit{Early Fusion} ($EF$), clinical features are concatenated with the imaging representation derived from \textit{Step 2} and processed via an MLP classifier. In \textit{Late Fusion} ($LF$), the imaging model (\textit{Step 2}) and the clinical model ($CL$) are trained independently, and their predictions are combined by averaging predicted probabilities. 

\textit{\textbf{Cross-Dataset Evaluation Protocol.}} 
To assess cross-institutional generalization, we adopt a leave-one-dataset-out strategy. Given its substantially larger size, I-SPY2 is always included in the training set, while the remaining datasets are alternately used as independent test sets, resulting in three separate scenarios: \textit{External-DUKE}, \textit{External-ISPY1}, and \textit{External-NACT}. 

The imaging backbone consists of the SwinUNETR~\cite{tang2022self} encoder, a transformer architecture pre-trained using a self-supervised learning strategy on large-scale 3D medical data~\cite{tang2022self}. 
Optimization was performed using the Adam optimizer and different learning rates were adopted across the stepwise training procedure to ensure stable convergence. In \textit{Step 1}, the model was trained with a learning rate of $5 \times 10^{-5}$ and weight decay of $1 \times 10^{-5}$. 
In \textit{Step 2} and \textit{Step 3}, the learning rate was reduced to $2 \times 10^{-5}$, while maintaining the same weight decay. 
For the MLP modules used in the $CL$ and in the $EF$ configuration, a learning rate of $1 \times 10^{-3}$ and weight decay of $1 \times 10^{-5}$ were adopted.
All configurations were trained for a maximum of 300 epochs, with an early stopping applied based on validation loss. 
For configurations using localization guidance, the hyperparameter $\lambda$ in \eqref{loss} was set to 100. This value was chosen after preliminary experiments exploring values from 0 to 200 (step size 50), providing the best trade-off between classification performance and spatial alignment. All models were implemented in
PyTorch and trained on a single NVIDIA A40 GPU.

To comprehensively evaluate model performance, we employed the following metrics: Specificity (Spe), Sensitivity (Sens), Balanced Accuracy (BA), and Area Under the Curve (AUC). BA is adopted to account for class imbalance in pCR prediction, ensuring a fair assessment across responder and non-responders.

\section{Results and Discussion}
\begin{figure}[t] 
    \centering 
    \includegraphics[width=0.92\textwidth]{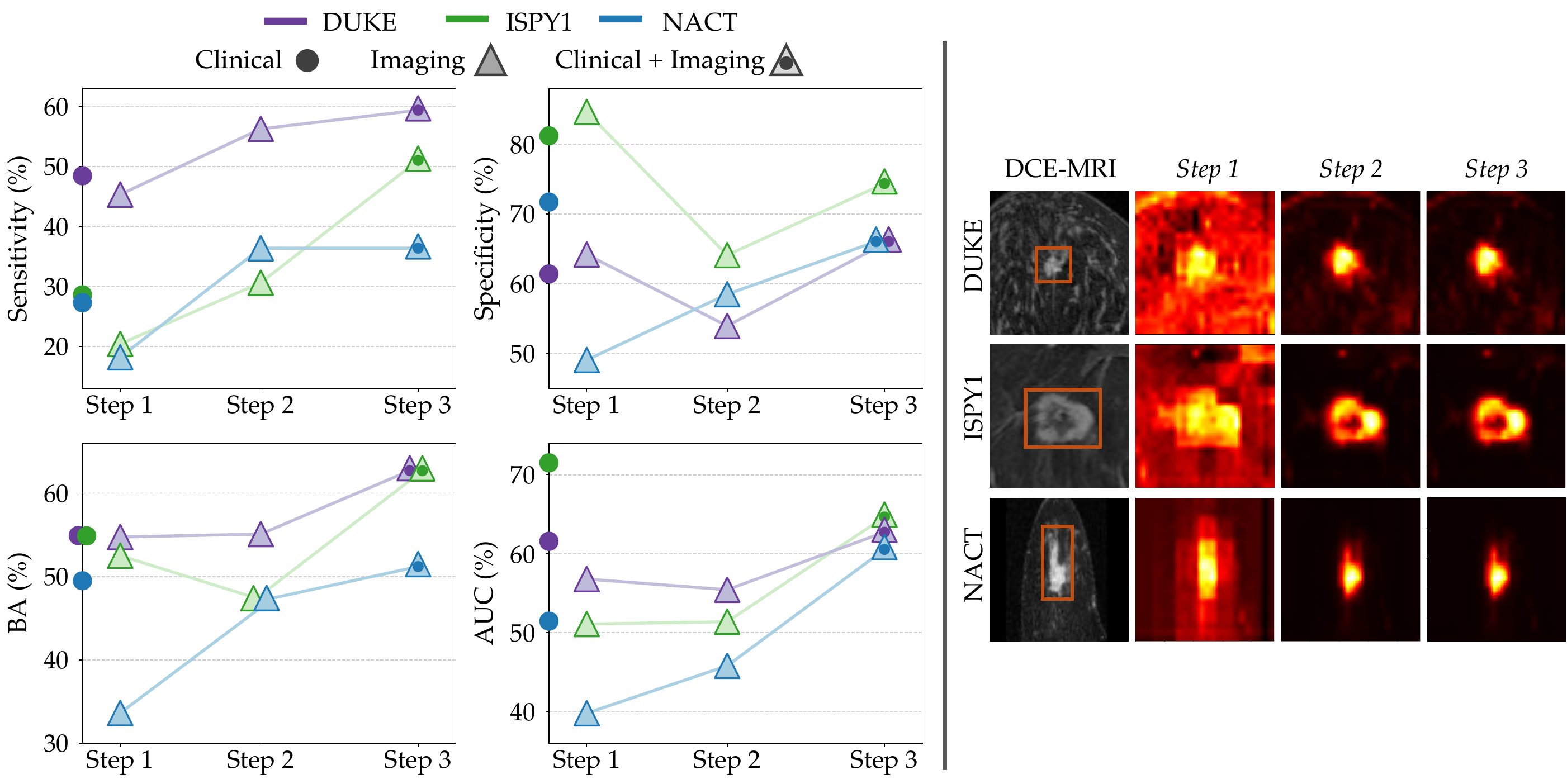} 
    \caption{{\textit{Left}: Quantitative performance across training steps under the external evaluation scenarios, with colors indicating the test cohort and marker shapes denoting modality configuration.
\textit{Right}: Attention maps across steps for representative patients from each cohort; bounding boxes highlight lesion regions on DCE-MRI.
    }}
    \label{fig:plot} 
\end{figure}

Table~\ref{tab:extended_table} summarizes the quantitative results across the three external evaluation scenarios, while Fig.~\ref{fig:plot} provides a visual overview of performance trends and corresponding attention maps across the stepwise framework.
\begin{table}[!hbt]
\centering
\caption{Performance metrics under the different experimental configurations. For each dataset and metric, the best-performing value is highlighted in \textbf{bold}. 
}
\label{tab:extended_table}
\resizebox{0.86\columnwidth}{!}{
\renewcommand{\arraystretch}{1.0}
\begin{tabular}{ccccccccccccc}
\toprule
 & \multicolumn{4}{c}{\textbf{External-DUKE}} & \multicolumn{4}{c}{\textbf{External-ISPY1}} & \multicolumn{4}{c}{\textbf{External-NACT}} \\
\cmidrule(lr){2-5}\cmidrule(lr){6-9}\cmidrule(lr){10-13}
\textbf{Approach} & Spe & Sens & BA & AUC  & Spe & Sens & BA & AUC  & Spe & Sens & BA & AUC \\
\midrule
{\textit{Step 1}} & 0.64 & 0.45 & 0.54 & 0.56 & \textbf{0.84} & 0.20 & 0.52 & 0.51 & 0.49 & 0.18 & 0.33 & 0.39 \\
{\textit{Step 2}} & 0.54 & 0.56 & 0.55 & 0.55 & 0.64 & 0.31 & 0.47 & 0.51 & 0.58 & \textbf{0.36} & 0.47 & 0.46 \\
{\textit{Step 3}} & \textbf{0.66} & \textbf{0.59} & \textbf{0.62} & \textbf{0.63} & 0.74 & \textbf{0.51} & \textbf{0.62} & 0.65 & 0.66 & \textbf{0.36} & \textbf{0.51} & \textbf{0.61} \\
\midrule
{$CL$} & 0.61 & 0.48 & 0.54 & 0.61 & 0.81 & 0.28 & 0.54 & \textbf{0.71} & 0.71  & 0.27 & 0.49 & 0.51 \\
\midrule
{$EF$} & \textbf{0.66} & 0.51 & 0.58 & 0.62 & 0.81 & {0.35} & 0.58 & 0.69 & {0.73} & 0.27 & 0.50 & 0.56 \\
\midrule
{$LF$} & \textbf{0.66} & 0.54 & 0.60 & 0.61 & 0.79 & 0.20 & 0.49 & 0.61 & \textbf{0.76} & 0.27 & \textbf{0.51} & 0.50 \\
\bottomrule
\end{tabular}
}
\end{table}

In \textit{External-DUKE}, the transition from \textit{Step 1} to \textit{Step 2} increases sensitivity, despite a reduction in specificity, reflecting a shift toward improved detection of responders. In \textit{Step 3}, specificity is recovered while sensitivity further improves, resulting in higher BA and AUC. This behavior indicates that the full framework enhances responder identification without compromising discrimination of non-responders, thereby achieving a more balanced predictive profile.

In \textit{External-ISPY1}, \textit{Step 1} shows high specificity but limited sensitivity. With \textit{Step 2}, sensitivity increases while specificity decreases, reflecting a rebalancing of decision boundaries. In \textit{Step 3}, sensitivity further improves and overall performance increases, as indicated by higher BA and AUC.
Although specificity does not exceed the initial peak (largely due to class imbalance), the final configuration achieves a more clinically meaningful sensitivity-specificity trade-off.

%In \textit{External-ISPY1}, \textit{Step 1} exhibits high specificity but limited sensitivity.
%In \textit{Step 2}, sensitivity increases, while specificity slightly drops, reflecting a transitional rebalancing of decision boundaries. In \textit{Step 3}, sensitivity increases substantially and overall performance improves, as reflected by higher BA and AUC. Although specificity does not surpass the initial peak, which is largely attributable to class imbalance, the final configuration achieves a more clinically meaningful trade-off, improving sensitivity while maintaining competitive specificity.

In the smaller \textit{External-NACT} cohort, overall performance is lower, yet the stepwise framework again demonstrates progressive gains, achieving the best performance in \textit{Step 3}. Notably, both specificity and sensitivity increase, supporting robustness of our approach even under limited data.

Across the unimodal models, $CL$ achieves similar or higher BA and AUC, but lower sensitivity.  \textit{Step 3} outperforms both configurations in terms of sensitivity and BA across all scenarios, yielding a more balanced predictive profile.

When compared with conventional fusion strategies, \textit{Step 3} achieves more balanced predictive behavior. Although $EF$ and $LF$ often achieve high specificity, this is accompanied by reduced sensitivity. Overall, the proposed approach proves more effective than feature concatenation or decision-level averaging.

The quantitative improvements observed across steps are accompanied by progressive spatial refinement, as shown by the attention maps in Fig.~\ref{fig:plot}. In \textit{Step 1}, attention is diffuse, reflecting reliance on global contextual patterns. With localization guidance in \textit{Step 2}, attention concentrates on the tumor core, and this anatomically coherent focus is preserved in \textit{Step 3}. 
Notably, even under localization guidance, attention is not always strictly confined to the mask:
activation may extend to surrounding tissue, suggesting exploitation of contextual cues related to pCR, while hypointense regions within the lesion may receive limited attention, indicating that the model selectively emphasizes discriminative imaging features rather than uniformly attending to the entire mask.

\section{Conclusion}
We introduced a multimodal stepwise clinically-guided attention learning framework for pCR prediction from breast DCE-MRI. By integrating tumor-driven spatial guidance and clinically modulated decision learning, the proposed strategy improves robustness across datasets and responder sensitivity under class imbalance while maintaining competitive specificity. It also generates anatomically coherent attention maps that provide insight into the model’s decision process.
These findings suggest that clinically grounded progressive attention learning is a promising direction for robust and transferable response prediction across heterogeneous clinical cohorts.

%\section{Note} 
% 1) valutare se inserire confronto con sota, 
% 2) dataset interni
% 3) riportare metriche generali sui dataset esterni
% 4) aggiungere ref

%% removed for anonymized MICCAI submission.

%\begin{credits}
%\subsubsection{\ackname} 
%A bold run-in heading in small font size at the end of the paper is used for general acknowledgments, for example: This study was funded by X (grant number Y).

%\subsubsection{\discintname}
%It is now necessary to declare any competing interests or to specifically state that the authors have no competing interests. Please place the statement with a bold run-in heading in small font size beneath the (optional) acknowledgments\footnote{If EquinOCS, our proceedings submission system, is used, then the disclaimer can be provided directly in the system.}, for example: The authors have no competing interests to declare that are relevant to the content of this article. Or: Author A has received research grants from Company W. Author B has received a speaker honorarium from Company X and owns stock in Company Y. Author C is a member of committee Z.
%\end{credits}

%giusto per vincolare le references ad iniziare da pagina 9 

\bibliographystyle{splncs04}
\bibliography{mybib}

\end{document}